\documentclass[lettersize,journal]{IEEEtran}
\usepackage{amsmath,amsfonts}
\usepackage{algorithmic}
\usepackage{algorithm}
\usepackage{array}
\usepackage[caption=false,font=normalsize,labelfont=sf,textfont=sf]{subfig}
\usepackage{textcomp}
\usepackage{stfloats}
\usepackage{url}
\usepackage{verbatim}
\usepackage{graphicx}
\usepackage{cite}

\usepackage{multirow}
\usepackage{booktabs}

\begin{document}

\title{Funnel-HOI: Top-Down Perception for Zero-Shot HOI Detection}

\author{Sandipan Sarma, Agney Talwarr, Arijit Sur
\thanks{Sandipan Sarma, Agney Talwarr, and Arijit Sur are with the Department of Computer Science and Engineering, Indian Institute of Technology, Guwahati, Assam 781039, India (e-mail: sandipan.sarma7@gmail.com; agneytalwar@gmail.com; arijit@iitg.ac.in).}
}

\markboth{Journal of \LaTeX\ Class Files,~Vol.~14, No.~8, August~2021}%
{Shell \MakeLowercase{\textit{et al.}}: A Sample Article Using IEEEtran.cls for IEEE Journals}


\maketitle

\begin{abstract}
Human-object interaction detection (HOID) refers to localizing interactive human-object pairs in images and identifying the interactions. Since there could be an exponential number of object-action combinations, labeled data is limited -- leading to a long-tail distribution problem. Recently, zero-shot learning emerged as a solution, with end-to-end transformer-based object detectors adapted for HOID becoming successful frameworks. However, their primary focus is designing improved decoders for learning entangled or disentangled interpretations of interactions. We advocate that HOI-specific cues must be anticipated at the encoder stage itself to obtain a stronger scene interpretation. Consequently, we build a top-down framework named Funnel-HOI inspired by the human tendency to grasp well-defined concepts first and then associate them with abstract concepts during scene understanding. We first probe an image for the presence of objects (well-defined concepts) and then probe for actions (abstract concepts) associated with them. A novel asymmetric co-attention mechanism mines these cues utilizing multimodal information (incorporating zero-shot capabilities) and yields stronger interaction representations at the encoder level. Furthermore, a novel loss is devised that considers object-action relatedness and regulates misclassification penalty better than existing loss functions for guiding the interaction classifier. Extensive experiments on the HICO-DET and V-COCO datasets across fully-supervised and six zero-shot settings reveal our state-of-the-art performance, with up to 12.4\% and 8.4\% gains for unseen and rare HOI categories, respectively.  
\end{abstract}

\begin{IEEEkeywords}
Zero-shot learning, human-object interaction detection, co-attention, transformer.
\end{IEEEkeywords}

\section{Introduction}
\label{sec:intro}

\IEEEPARstart{G}{iven} an image, the objective in human-object interaction detection (HOID) is to localize human-object pairs involved in an interaction and recognize the interaction categories represented as a $\langle subject, verb, object
 \rangle$ triplet, e.g. $\langle human, ride, horse \rangle$. However, most HOID models today are supervised learners. With new interactions emerging frequently in this era of social media and surveillance systems, collecting annotated data for these novel human-object interaction (HOI) categories is costly, resulting in a long-tailed distribution for HOI classes. Consequently, it compromises their ability to detect unseen HOIs in the wild.   

\begin{figure}[t]
  \centering
  \includegraphics[width=\linewidth]{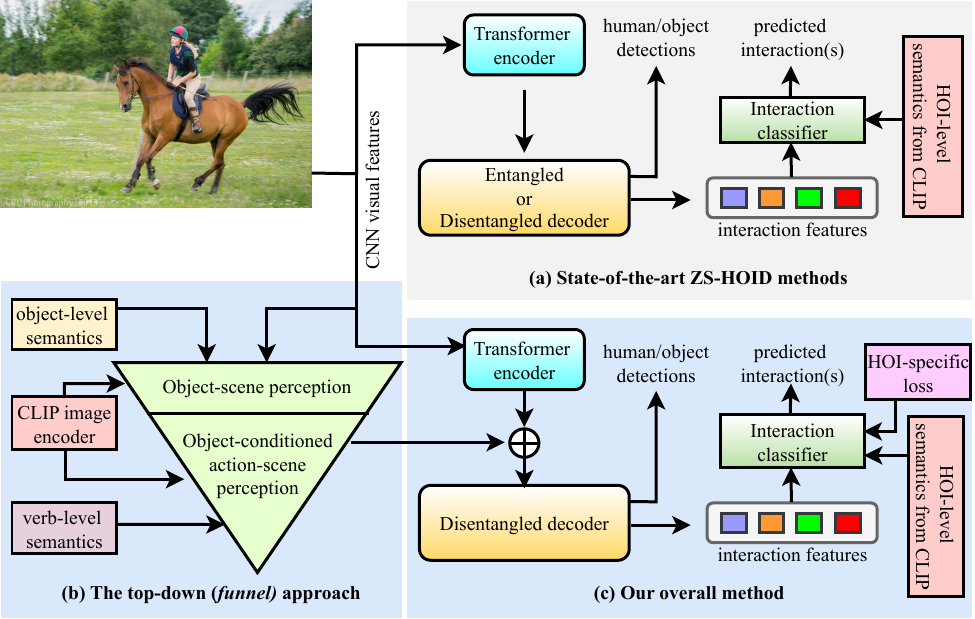}
  \caption{Comparison between DETR-based zero-shot HOID methods. (a) Current methods lack object and verb-specific cues while capturing the scene using encoders. (b) We integrate multimodal information -- first for the object and then for the associated verbs, imitating the human tendency of top-down scene understanding. (c) Our HOI-specific loss function improves the overall framework.}
  \label{fig:teaser}
\end{figure}

Zero-shot HOID (ZS-HOID) was proposed in~\cite{shen2018scaling} as a solution to this data scarcity problem, where a model learns to detect unseen HOIs (having no visual examples during training) at test time by transferring its knowledge about \textit{seen} HOIs to the \textit{unseen} as per visual-semantic associations. Recently, following the success of DETR~\cite{carion2020end} as an impressive end-to-end object detector, ZS-HOID methods have modified and adapted it for interaction detection, witnessing significant boosts in performance. However, there remain three major limitations.

First, all the DETR-based methods are primarily decoder-centric (Fig.~\ref{fig:teaser}a). The initial proposal by~\cite{liao2022gen} was to disentangle the ZS-HOID task into two steps -- human-object pair localization and interaction classification -- using two separate decoders. Their interaction decoder is refined in ~\cite{wu2023end} by designing a two-stage bipartite matching algorithm and training an interactive score module. HOICLIP~\cite{ning2023hoiclip} infuses additional guidance via CLIP~\cite{radford2021learning} into the interaction decoder from a pretrained visual-language model using a cross-attention transformer. Among all these efforts to improve the quality of interaction features given by the decoder, the encoder remains largely ignored. The vanilla transformer encoder captures image-wide contextual information but is not designed to mine HOI-specific cues, which could help the decoders adapt better to ZS-HOID. Second, existing approaches use semantic information only at the HOI classification step at the end. Despite visual-semantic association being the core principle in zero-shot learning, the sole reliance on visual features for training the encoder and decoders inhibits the potential of the DETR framework for ZS-HOID. And third, the focal loss commonly used in ZS-HOID is a straightforward adaptation of the one devised originally for object detection~\cite{lin2017focal} and does not consider several factors in a complicated task like HOID, such as relationships between objects, verbs, and HOI classes. Moreover, it imposes a high penalty on misclassifications only for pulling the HOI predictions closer to ground truths, without inherently pushing them apart from the irrelevant classes.

To resolve the aforementioned issues, we make three-fold contributions in this paper. Motivated by psychological studies that primates and humans associate actions with object-compatibility~\cite{schubotz2014objects}, we design a novel co-attention mechanism in a top-down manner~\cite{gaspelin2018top} (hence the name \textbf{\textit{Funnel}}-HOI). The idea, as shown in Fig.~\ref{fig:teaser}b, is to perceive well-defined concepts (objects) first and then capitalize on them to grasp more specific, abstract concepts (actions). Specifically, our proposed \textbf{Object-Specific Asymmetric Co-Attention (OSACA)} module first leverages multimodal (visual-semantic) inputs to produce an object-probed feature map, indicating object presence in the image. In the next step, actions highly associated with the objects perceived from the previous step are ascertained. Then, we train an \textbf{Object-conditional Verb-specific Asymmetric Co-Attention (OVACA)} module with multimodal inputs guided by the object-probed feature map. The output is a verb-probed feature map which is used to enhance the vanilla encoder output (Fig.~\ref{fig:teaser}c). Meanwhile, we also propose a novel \textbf{Object-Regulated Discrepancy (ORDis) loss}  which analyzes HOI-specific cues from the scene and addresses the limitations of the conventional focal loss~\cite{lin2017focal}. Experiments conducted on the challenging HICO-DET dataset~\cite{chao2018learning} in six different zero-shot settings demonstrate that ZS-HOID benefits from having multimodal knowledge within the detection framework instead of only at the final classification step. Furthermore, our attention visualizations highlight the effectiveness of the co-attention modules and the ORDis loss. We also demonstrate our method's robustness by presenting results on HICO-DET and V-COCO~\cite{gupta2015visual} datasets in the fully-supervised setting.  To summarize, our contributions are as follows:

\begin{itemize}
    \item We propose \textit{Funnel-HOI}, a ZS-HOID approach to obtaining multimodal HOI features from an image. A novel co-attention mechanism is proposed to probe for the presence of objects highly relevant to the image. 
    \item With a perception of the relevant objects, we propose a verb nomination strategy. The proposed co-attention is again applied for probing verbs but with the object-probed feature map as an input condition.
    \item Object and verb-level semantics (Fig.~\ref{fig:teaser}b) are utilized \textit{separately} for the first time in ZS-HOID understanding. 
    \item A novel ORDis loss function is designed to acknowledge HOI-specific factors for improved interaction detection.

\end{itemize}

\begin{figure*}[t]
  \centering
  \includegraphics[width=\linewidth]{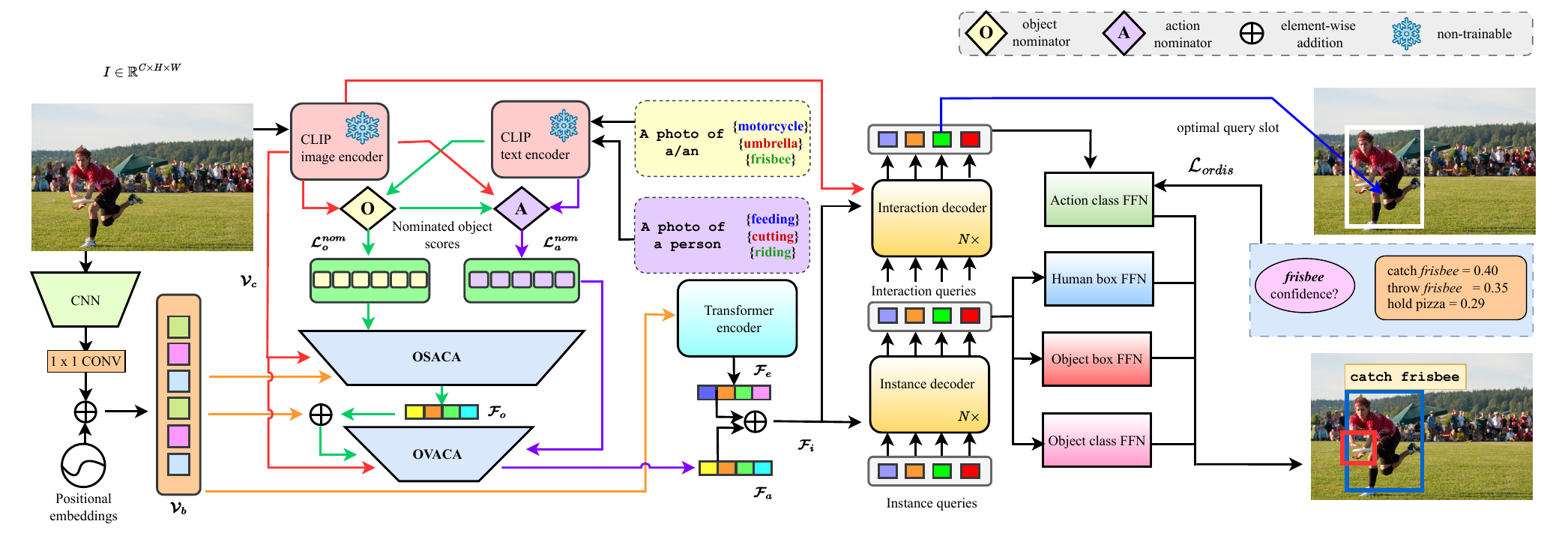}
  \caption{The Funnel-HOI pipeline: Given an image, visual features are extracted from CNN backbone ($\boldsymbol{\mathcal{V}_b}$) and CLIP image encoder ($\boldsymbol{\mathcal{V}_c}$). A few objects relevant to the image are nominated, for which OSACA produces an object-probed feature map $\boldsymbol{\mathcal{F}_o}$ (top-level understanding). Next, actions that are highly associated with the nominated objects are picked for which OVACA computes a verb-probed feature map $\boldsymbol{\mathcal{F}_a}$ (deeper understanding). The interaction encodings obtained ($\boldsymbol{\mathcal{F}_i}$) are fed to the decoders. For HOI-specific optimization, the novel ORDis loss examines object-verb relationships. Best viewed in color.}
  \label{fig:archi}

\end{figure*}

\section{Related work}
 \label{sec:lit}

 \subsection{Human-Object Interaction Detection (HOID)}
 \label{sec:lit-hoid}
 Existing literature is broadly divided into two-stage and one-stage detection methods. Two-stage methods~\cite{gkioxari2018detecting,gao2018ican,xu2019interact, hou2020visual, gao2020drg, hou2021affordance, gao2021hierarchical, wang2021ipgn, liao2024ppdm++, wu2024toward} localize the humans and objects in the image first using pretrained object detectors~\cite{ren2015faster} and then process visual features extracted from the regions of interest via multi-stream networks. Some of them focus on object-verb connectivity like us -- for instance, ACP++~\cite{kim2021acp++} looks for co-occurrences between interactions (formed by object and verb together). However, our hierarchical approach is different as we disentangle object and verb dependencies.
 
 Meanwhile, one-stage methods came to the limelight, especially after the advent of DETR~\cite{carion2020end}, achieving end-to-end HOI detections. Many recent works~\cite{tamura2021qpic,zhang2021mining,zhang2022efficient,ma2023fgahoi, zong2023zero, chan2024region, cheng2024parallel} have adopted DETR for detection and modified its decoder additionally for human-object pair detection and interaction classification. Nevertheless, works like these are predominantly supervised and fail in zero-shot scenarios.

\subsection{Zero-shot HOID (ZS-HOID)}
\label{sec:lit-ZS-HOID}
Since their inception, ZS-HOID methods have broadly followed three research directions. The earliest were two-stage ZS-HOID methods, with contributions mostly in the second stage. 
To this end, the unseen interactions, objects, and verbs are conceptualized in several ways, including using outputs from separate verb and object networks~\cite{shen2018scaling}, generating HOI triplets for functionally-similar objects~\cite{bansal2020detecting}, and learning consistency graphs~\cite{liu2020consnet}. Another approach was to instinctively pose the HOID task as a compositional learning problem~\cite{hou2020visual,hou2021detecting,hou2022discovering}. 
The recent approaches reduce model complexity and dependency on external knowledge using DETR-based frameworks. GEN-VLKT~\cite{liao2022gen} is the first work in this direction that uses disentangled decoders and strengthens the interaction classifier using CLIP textual embeddings~\cite{radford2021learning}. HOICLIP~\cite{ning2023hoiclip} uses CLIP visual embeddings as a guide to enhance the interaction decoder via a cross-attention mechanism. EoID~\cite{wu2023end} additionally enables ``potential interaction'' detection by devising a two-stage bipartite matching algorithm.
However, the inputs to these decoders come from the vanilla transformer encoder, which does not explicitly model object-verb relationships. In addition, the textual modality remains underutilized -- used only 
during final interaction classification step. We leverage object and verb-level semantics to strengthen encoder representations.

Regarding loss function for interaction classification, initial works utilized the cross-entropy loss, or its modified versions for estimating HOI prediction error~\cite{shen2018scaling,bansal2020detecting,hou2020visual,hou2021detecting,hou2021affordance}. In contrast, the DETR-based frameworks prefer using a focal loss~\cite{lin2017focal} for training HOI classifiers~\cite{liao2022gen,wu2023end,ning2023hoiclip}. Although an affordance graph-based regularization loss is proposed~\cite{sarullo2020zero} to group functionally-similar actions, it does not jointly consider the impact of objects, which we do while devising our loss. 

\section{Approach}
\label{sec:approach}

\subsection{Problem formulation}
\label{sec:problem}
In this work, we focus on ZS-HOID in images. Let the dataset contain a set of $C$ HOI classes. We define $\mathcal{Y}^s = \{ y_1, y_2,... y_S \}$ and $\mathcal{Y}^u = \{ y_{(S+1)}, y_{(S+2)},...y_{(S+U)} \}$ as the label sets of $S$ number of seen and $U$ number of unseen HOI classes respectively, such that $ C = S+U $ and $\mathcal{Y}^s \cap \mathcal{Y}^u = \phi$. An \textit{interaction} is described as a triplet $\langle b_h, b_o, y_i \rangle$ where
$b_h, b_o \in \mathbb{R}^{4}$ indicates the bounding box coordinates of the interacting human and object respectively, and $y_i \in \mathcal{Y} = \{ y_1, y_2,...y_C \}$ denotes an HOI class label. Every HOI class $y_i$ is formed by combining an action class 
$a \in \mathcal{A} = \{a_1, a_2..., a_M\}$ with an object class 
$o \in \mathcal{O} = \{o_1, o_2,..., o_N\}$, so $y = \langle a, o \rangle$. Then, the training data becomes:
\begin{align}
    \label{eq:train_data}
    \mathcal{D}^{train} = \{I_k, \{H^k_p \}_{p = 1}^{N_h}, \{O^k_q \}_{q = 1}^{N_o}, \{y^k_r \}_{r = 1}^{N_i} \}_{k = 1}^{K}
\end{align}
 that contains $K$ images, having $N_h$ humans, $N_o$ objects, and $N_i$ interactions, where a human may interact with multiple objects in a single image. For an image $I_k$, the $p^{th}$ human is annotated as $H^k_p = \{ b_h, o_{person} \}$ and every $q^{th}$ object as $O^k_q = \{ b_o, o_{obj} \}$, where $o_{person}, o_{obj} \in \mathcal{O}$ is the human/object class label. Moreover, only seen interaction labels are accessible during zero-shot training, i.e., $y^k_r \in \mathcal{Y}^s$. With this annotated data at hand, the task in ZS-HOID becomes learning a model that takes images containing HOIs from both $\mathcal{Y}^s$ and $\mathcal{Y}^u$ at test time and detecting the interacting human-object pairs while predicting the interaction labels. With these definitions, we now describe Funnel-HOI in the following sections.

 \subsection{Object-Specific Asymmetric Co-Attention (OSACA)}
 \label{sec:osaca}

\subsubsection{\textbf{Object nominator}}
\label{sec:nomo}
In our top-down approach to scene understanding, the first step is to probe for object presence using joint visual-textual cues. Since an object is a \textit{well-defined concept} characterized by unique attributes, we extract the semantics (textual cues) of all objects by first converting an object name to ``\texttt{A photo of a/an [OBJ]}'' via a prompt template $\rho_o(.)$, and then using a pretrained CLIP text encoder~\cite{radford2021learning} $\psi(.)$ as: 
\begin{align}
    \label{eq:lo}
    \boldsymbol{\mathcal{L}_o} = \left[ \boldsymbol{l_o^1}, \boldsymbol{l_o^2},...\boldsymbol{l_o^N} \right] \in \mathbb{R}^{d \times N}, \quad \boldsymbol{l_o^i} = \psi(\rho_o(o_i)) \quad \forall \,o_i \in \mathcal{O}
\end{align}
However, the images of unseen interactions/objects are unavailable during zero-shot training, and hence, we cannot use scores from pretrained supervised classifiers to recognize an object. Instead, we rely on the visual features of the scene $\boldsymbol{\mathcal{V}_c} \in \mathbb{R}^d$ obtained from the image encoder of CLIP. An object-scene similarity is obtained:
\begin{align}
    \label{eq:nomo}
    S_{os} =  \boldsymbol{\mathcal{L}_o}^T \boldsymbol{\mathcal{V}_c}
\end{align}
and the top-$K_o$ objects from the object set $\mathcal{O}$ are nominated as the best candidates that summarize the scene. Moreover, since humans are always present in a HOID task, we explicitly add the ``person" class to the nomination list and obtain $\boldsymbol{\mathcal{L}_o^{nom}} \in \mathbb{R}^{d \times (K_o + 1)}$ as a submatrix of $\boldsymbol{\mathcal{L}_o}$.

\subsubsection{\textbf{Asymmetric co-attention}}
\label{sec:o-coatt}
The intuition behind using co-attention is to jointly learn how visual understanding is influenced by textual knowledge about the few nominated objects and vice-versa for best describing an image $I \in \mathbb{R}^{C \times H \times W}$.
To this end, we propose a novel co-attention mechanism, with powerful CLIP visual features $\boldsymbol{\mathcal{V}_c}$ (reshaped to $\mathbb{R}^{C_1 \times 2}$) acting as a guide. With CNN-visual features of $I$ ($\boldsymbol{\mathcal{V}_b} \in \mathbb{R}^{C_1 \times H'W'}$) and reshaped textual features $\boldsymbol{\mathcal{L}_o^{nom}} \in \mathbb{R}^{C_1 \times (2 \times N_{obj})}$, their affinity matrix is:
\begin{align}
    \label{eq:affinity}
    \boldsymbol{\Gamma} = (\boldsymbol{W_v}^{T} \boldsymbol{\mathcal{V}_b})^T (\boldsymbol{W_l}^{T} \boldsymbol{\mathcal{L}_o^{nom}}) 
\end{align}
where $\boldsymbol{W_v}, \boldsymbol{W_l} \in \mathbb{R}^{C_1 \times C_2}$ are learnable weights, $C_2 < C_1$, and we redefine $N_{obj} = K_o + 1$. Note that the feature reshapings have been done to ensure compatible matrix multiplications. Unlike vanilla co-attention where the affinity matrix is square~\cite{li-etal-2018-co, lu2019see}, ours is rectangular ($\boldsymbol{\Gamma} \in \mathbb{R}^{H'W' \times (2 \times N_{obj})}$) as it learns the affinity between a single visual scene and multiple \textit{object-specific} semantics simultaneously. Therefore, we call it \textit{\textbf{asymmetric}} co-attention. The affinity matrix is used as a feature to learn visual-semantic entanglement via further attention maps. As shown in Fig.~\ref{fig:aca}, we first explore how strongly the visual modality is influenced by the affinity matrix as follows:
\begin{figure}[t]
  \centering
  \includegraphics[width=0.8\linewidth]{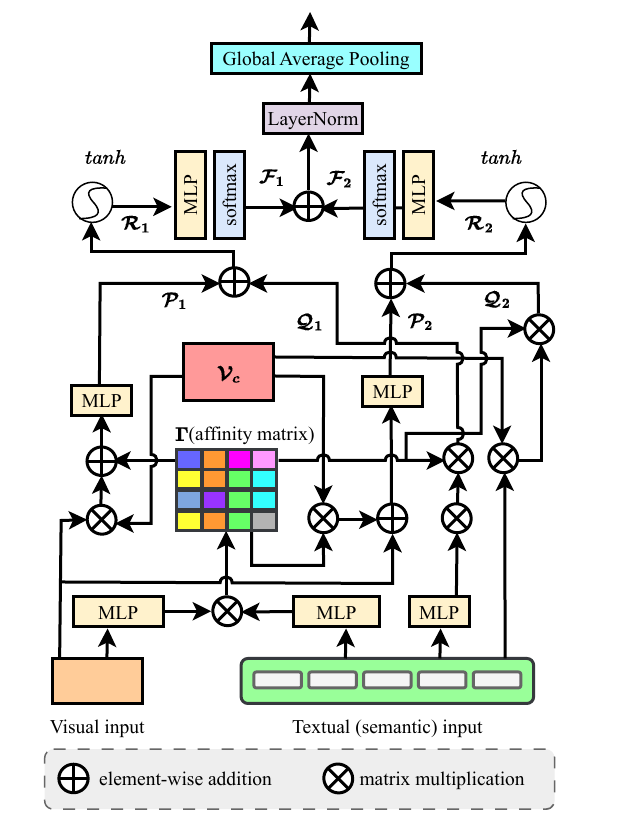}
  \caption{Structure of asymmetric co-attention for both OSACA and OVACA. The difference lies in the inputs. The visual input is just the CNN features for OSACA but is added with the object-probed feature map for OVACA (Eq.~\ref{eq:v+fo}). Semantic vectors of the nominated objects and verbs are the textual inputs in OSACA and OVACA, respectively. }
  \label{fig:aca}
\end{figure}
\begin{align}
    \label{eq:t1v}
    \boldsymbol{\mathcal{P}_1} &= \left( \left[ \boldsymbol{\mathcal{V}_b}^T \boldsymbol{\mathcal{V}_c} \right]_{N_{obj}} + \boldsymbol{\Gamma} \right) \boldsymbol{W_1} \\
    \label{eq:t2l}
    \boldsymbol{\mathcal{P}_2} &= \left( \left[ \boldsymbol{\mathcal{V}_b}^T \right]_{N_{obj}} + \boldsymbol{\Gamma} \left[ \boldsymbol{\mathcal{V}_c}^T \right]_{N_{obj}} \right) \boldsymbol{W_2}  
\end{align}
where $\left[ x \right]_{N_{obj}}$ indicates $x$ is duplicated $N_{obj}$ times and concatenated. Next, the influence of $\boldsymbol{\Gamma}$ on textual modality is inferred as:
\begin{align}
    \label{eq:t2v}
    \boldsymbol{\mathcal{Q}_1} &= \boldsymbol{\Gamma}(\boldsymbol{\mathcal{L}_o^{nom}}^T \boldsymbol{W_3}) \\
    \label{eq:t1l}
    \boldsymbol{\mathcal{Q}_2} &=((\boldsymbol{\mathcal{V}_c}^T \boldsymbol{\mathcal{L}_o^{nom}})^T \boldsymbol{\Gamma}) \, \boldsymbol{W_4} 
\end{align}
The resultant feature maps are fused for summarization:
\begin{align}
    \label{eq:av}
    \boldsymbol{\mathcal{R}_1} = tanh(\boldsymbol{\mathcal{P}_1} + \boldsymbol{\mathcal{Q}_1}), \; \boldsymbol{\mathcal{F}_1} = softmax(\boldsymbol{W_5} \boldsymbol{\mathcal{R}_1}^T) \\
    \label{eq:al}
    \boldsymbol{\mathcal{R}_2} = tanh(\boldsymbol{\mathcal{P}_2} + \boldsymbol{\mathcal{Q}_2}), \; \boldsymbol{\mathcal{F}_2} = softmax(\boldsymbol{W_6} \boldsymbol{\mathcal{R}_2}^T)
\end{align}
These attention maps computed corresponding to all the nominated objects are summarized using global average pooling (GAP), yielding an object-probed feature map $ \boldsymbol{\mathcal{F}_o}$:
\begin{align}
    \label{eq:fo}
    \boldsymbol{\mathcal{F}_o} = GAP(LN(\boldsymbol{\mathcal{F}_1} + \boldsymbol{\mathcal{F}_2}))
\end{align}
where $LN(.)$ denotes the layer normalization operation. $\boldsymbol{\mathcal{F}_o}$ indicates object presence, and in the next section, we discuss how it is used to dictate action interpretation in the image.

\subsection{Object-conditional Verb-specific Asymmetric Co-Attention (OVACA)}
\label{sec:ovaca}

\subsubsection{\textbf{Action nominator}}
\label{sec:noma}
Once we anticipate object presence, the next step of our top-down ``funnel'' approach is to leverage this information and probe for verbs (or actions). However, unlike objects, verbs are not well-defined and exhibit polysemy~\cite{liu2020consnet, zhong2020polysemy}, resulting in different human poses and appearances based on the context of the interactive object. To mimic this object-dependency, we first obtain the Object-Verb Relatedness (\textbf{OVeR}) matrix as follows:
\begin{align}
    \label{eq:over}
    \boldsymbol{\Delta} &= \boldsymbol{\mathcal{L}_o}^T \boldsymbol{\mathcal{L}_a} \in \mathbb{R}^{N \times M}, \quad \text{where:} \\
    \label{eq:la}
    \boldsymbol{\mathcal{L}_a} &= \left[ \boldsymbol{l_a^1}, \boldsymbol{l_a^2},...\boldsymbol{l_a^M} \right] \in \mathbb{R}^{d \times M}, \; \boldsymbol{l_a^i} = \psi(\rho_a(a_i)) \; \forall \,a_i \in \mathcal{A}
\end{align}
Here, $\rho_a(.)$ is the prompt: ``\texttt{A photo of a person [VERB]-ing}" and $[\Delta_{ij}] \in  [-1, 1]$. For each of the $K_o$ nominated objects, textual features of the top-$K$ related verbs (as per OVeR) are selected ($\boldsymbol{\mathcal{L'}_a} \in \mathbb{R}^{d \times K_o \times K}$) to find their alignment with the visual features of CLIP. Thus, we compute action-scene scores as:
\begin{align}
    \label{eq:noma}
    S_{as} = exp(\left[ S_{os}^{nom} \right]_K) \circ (\boldsymbol{\mathcal{V}_c}\boldsymbol{\mathcal{L'}_a})
\end{align}
where $\circ$ denotes Hadamard product and $S_{os}^{nom}$ contains the object-scene scores for the nominated objects. These exponentiated scores quantify the confidence of object presence in the image and regulate how likely an action is.
The top-$K_a$ verbs as per $S_{as}$ are nominated as the best candidates representing the actions in the image. Consequently, we obtain $\boldsymbol{\mathcal{L}_a^{nom}} \in \mathbb{R}^{d \times K_a}$ as a submatrix of $\boldsymbol{\mathcal{L}_a}$.

\subsubsection{\textbf{Object-conditional co-attention}}
\label{sec:v-coatt}
Since actions rely heavily on the objects present in the scene, we use the co-attention feature map $\boldsymbol{\mathcal{F}_o}$ as conditional information:
\begin{align}
    \label{eq:v+fo}
    \boldsymbol{\mathcal{V}_{oc}} = \boldsymbol{\mathcal{V}_b} + \boldsymbol{\mathcal{F}_o}
\end{align}
Using $\boldsymbol{\mathcal{V}_{oc}}$ and $\boldsymbol{\mathcal{L}_{a}^{nom}}$ as verb-specific inputs, we construct an asymmetric co-attention mechanism in the same fashion as described in Sec.~\ref{sec:o-coatt}. The output verb-probed feature map $\boldsymbol{\mathcal{F}_a}$ represents a top-down feature representation of the scene as a whole -- forming a conception about objects, using it to infer actions and yielding interaction encodings. Our final interaction encodings are:
\begin{align}
    \label{eq:enc_output}
    \boldsymbol{\mathcal{F}_i} = \boldsymbol{\mathcal{F}_a} + \boldsymbol{\mathcal{F}_e}
\end{align}
where $\boldsymbol{\mathcal{F}_e}$ denotes output from vanilla transformer encoder. Following HOICLIP~\cite{ning2023hoiclip}, our interaction encodings are first processed by an instance decoder and then passed through an interaction decoder that produces interaction embeddings. An interaction classifier is trained with these embeddings to predict the HOI class label and several other predictions, as shown in Fig.~\ref{fig:archi}.

\subsection{Object-Regulated Discrepancy (ORDis) loss}
\label{sec:ordis}
 Recent DETR-based HOID methods~\cite{liao2022gen, wu2023end, ning2023hoiclip} have had huge success using focal loss~\cite{lin2017focal} to train the interaction classifier. The improved performance can be attributed to its ability to deal with class imbalance, which is common in HOID. However, in a zero-shot scenario, predictions often become biased toward the seen classes during testing. Hence, in addition to ensuring that predictions corresponding to the interaction embeddings are close to the ground truth, pushing them away from the other classes also becomes crucial for updating the network and learning discriminative features. Besides, while adopting an object detector like DETR for ZS-HOID, existing works have not proposed any HOI-specific loss function. In the following sections, we first provide an insight into how DETR-based detection works and then propose HOI-aligned adjustments to focal loss.

\subsubsection{\textbf{Loss computation in DETR-based HOID}}
\label{sec:detr}
For an image $I$, DETR probes for objects within $N_q$ different \textit{query slots} ($\mathcal{Z} = \{ z_1, z_2,...z_{N_q}\}$) which learn to specialize on certain areas of the image, with several box sizes and operating modes~\cite{carion2020end}. A ZS-HOID design considers that the interaction embedding corresponding to each slot is associated with at most one human-object pair~\cite{tamura2021qpic, liao2022gen}. For loss computation, the first step is to use the output interaction embeddings $\boldsymbol{\mathcal{F}_d} \in \mathbb{R}^{N_q \times d}$ from the interaction decoder for HOI prediction and find a bipartite matching between slot predictions and ground truths via Hungarian algorithm~\cite{kuhn1955hungarian}. In the ZS-HOID context, it optimizes various costs~\cite{liao2022gen} simultaneously, say with a total \textit{optimal cost} of $\eta_I$. In the second step, the matched pairs are used for loss calculation~\cite{liao2022gen}, among which our focus will be particularly on focal loss for interaction classification. We call a query slot matching a ground truth as \textit{optimal slot}, with optimal slot cost $\eta (z_g)$ during Hungarian cost assignment such that $\eta_I = \sum_{g=1}^{G} \, \eta (z_g)$ (corresponding to $G$ ground-truth HOIs in image $I$).

\subsubsection{\textbf{The regulating factor}}
\label{sec:beta}
Each optimal query slot $z_i$ has an associated vector of prediction probabilities $\hat{z}_i \in \mathbb{R}^S$ during zero-shot training. Our goal is to tune the penalty for negative predictions (i.e., misclassifications) by focusing on hard examples~\cite{lin2017focal} (like focal loss), but provided that we possess a perception about the interactive object. Since each optimal slot is associated with a single object $o_i$ (due to bipartite matching) and optimal slot cost $\eta (z_i)$, it can be said that a higher $\eta (z_i)$ indicates lower confidence about associating $o_i$ to $z_i$, and vice-versa. Consequently, we propose:
\begin{align}
    \label{eq:beta}
    \beta (z_i) = \,
    \begin{cases}
        \log (1 + (\eta (z_i) + \epsilon_1)^{-\kappa}), &\text{if } z_i \text{ is an optimal slot} \\
        0, &\text{otherwise}
    \end{cases}
\end{align}
This factor regulates object perception in our proposed loss, where $\kappa$ and $\epsilon_1$ are constants. 

\subsubsection{\textbf{The relatedness factor}}
\label{sec:delta}
Once we understand object perception for slot $z_i$, the relatedness of the associated object $o_i$ to different verbs should also be accounted for while penalizing a negative HOI prediction. To this end, we use the OVeR matrix (Eq.~\ref{eq:over}) to define the relatedness factor between the object $o_i$ (associated to slot $z_i$) and an action $a_j$ for a valid seen class interaction label as:
\begin{equation}
    \label{eq:delta}
    \delta (z_i) = \quad
    \begin{cases}
        \Delta_{ij}, &\text{if } z_i \text{ is an optimal slot and} \, \langle o_i, a_j \rangle \in \mathcal{Y}^s \\
        0, &\text{otherwise}
    \end{cases}
\end{equation}

\begin{table}[t]
    \caption{Zero-shot performance comparison on HICO-DET. The best and second-best results are bold and underlined.}
    \label{tab:results}
    \centering
    \begin{tabular}{lcccc}
    
    \toprule
    \textbf{Method} & \textbf{Type} & \textbf{Unseen} & \textbf{Seen} & \textbf{Full} \\
    \midrule
    Functional~\cite{bansal2020detecting} & UC & 11.31$\pm$1.03 & 12.74$\pm$0.34 &  12.45$\pm$0.16 \\
    ConsNet~\cite{liu2020consnet} & UC & 16.99$\pm$1.67 & 20.51$\pm$0.62 &  19.81$\pm$0.32\\
    GEN-VLKT\cite{liao2022gen} & UC & 20.64$\pm$0.89 & 27.16$\pm$0.88 &  25.23$\pm$0.59\\
    EoID~\cite{wu2023end} & UC & \underline{23.01$\pm$1.54} & \underline{30.39$\pm$0.40} &  \underline{28.91 $\pm$ 0.27} \\
    \midrule
    \textbf{Funnel-HOI} & $\mathrm{UC}$ & \textbf{24.50$\pm$1.39} & \textbf{31.89$\pm$0.82} & \textbf{30.41$\pm$0.47} \\
    \midrule
    VCL~\cite{hou2020visual} & RF-UC & 10.06 & 24.28 & 21.43 \\

    FCL~\cite{hou2021detecting}  & RF-UC & 13.16 & 24.23 & 22.01 \\
    SCL~\cite{hou2022discovering} & RF-UC & 19.07 & 30.39 & 28.08 \\
    GEN-VLKT~\cite{liao2022gen} & RF-UC & 21.36 & 32.91 & 30.56 \\
    OSHOI~\cite{wu2024toward} &RF-UC &23.32 &30.09 &28.53 \\
    EoID~\cite{wu2023end} & RF-UC & 22.04 & 31.39 & 29.52 \\
    HOICLIP~\cite{ning2023hoiclip} & RF-UC & \underline{25.52} &  \textbf{34.72} &  \textbf{32.88} \\
    \midrule
    \textbf{Funnel-HOI} & RF-UC &  \textbf{25.90} & \underline{34.43} & \underline{32.70} \\
    \midrule
  
    FCL~\cite{hou2021detecting}  & NF-UC & 18.66 & 19.55 & 19.37 \\
    SCL~\cite{hou2022discovering} & NF-UC & 21.73 & 25.00 & 24.34 \\ 
    GEN-VLKT~\cite{liao2022gen}  & NF-UC & 25.05 & 23.38 & 23.71 \\
    OSHOI~\cite{wu2024toward} &NF-UC & \textbf{27.35} & 22.09 &23.14 \\
    EoID~\cite{wu2023end} & NF-UC &  \underline{26.77} & 26.66 & 26.69 \\
    HOICLIP~\cite{ning2023hoiclip} & NF-UC & 26.39 & \underline{28.10} & \underline{27.75} \\
    \midrule
    \textbf{Funnel-HOI} & NF-UC & 26.63 &  \textbf{28.32} &  \textbf{28.00} \\
    \midrule

    Functional~\cite{bansal2020detecting} & UO & 11.22 & 14.36 &  13.84 \\

    FCL~\cite{hou2021detecting}  & UO & 0.00 & 13.71 & 11.43 \\
    GEN-VLKT~\cite{liao2022gen} & UO & 10.51 & 28.92 & 25.63 \\

    HOICLIP~\cite{ning2023hoiclip} & UO &  \underline{13.59} & \underline{30.99} & \underline{28.53} \\
    \midrule
    \textbf{Funnel-HOI} & UO & \textbf{15.28} &  \textbf{31.50} &  \textbf{28.66} \\
    \midrule
    
    ConsNet~\cite{liu2020consnet}  & UA & 12.50 & 14.72 & 14.35 \\
    CDT~\cite{zong2023zero} & UA & 15.17  & 21.45 & 19.68\\
    EoID~\cite{wu2023end} & UA & \underline{23.04}  & \underline{30.46} & \underline{29.22}\\
    OSHOI~\cite{wu2024toward} & UA & 17.92 &28.13 &26.43 \\
    \midrule
    \textbf{Funnel-HOI} & UA &  \textbf{24.11} &  \textbf{31.85} &  \textbf{30.56} \\
    \midrule
    
    GEN-VLKT~\cite{liao2022gen} & UV & 20.96 & 30.23 & 28.74 \\
    EoID~\cite{wu2023end} & UV & 22.71  & 30.73 & 29.61\\
    HOICLIP~\cite{ning2023hoiclip} & UV & \underline{24.30} & \underline{32.19} & \underline{31.09} \\
    \midrule
    \textbf{Funnel-HOI} & UV &  \textbf{26.10} &  \textbf{32.56} &  \textbf{31.66} \\
    \bottomrule
    \end{tabular}

\end{table}

\subsubsection{\textbf{The inter-class discrepancy factor}}
\label{sec:zeta}
The vector $\hat{z}_i$ matched to ground-truth $y_i$ contains $S$ number of predictions -- some correspond to ground-truth/positive classes (say a set $\hat{z}_i^+$), and others correspond to negative classes (say a set $\hat{z}_i^-$). Note that there can be multiple HOIs associated with a single human-object pair. Prediction scores within $\hat{z}_i^+$ should be ideally high, and those in $\hat{z}_i^-$ should be low. However, when the model performs poorly during training, it is vital to direct the training properly by learning from the inter-class prediction discrepancy. Specifically, if the predicted probability of a negative class exceeds the average predicted probability from $\hat{z}_i^+$, it implies a large discrepancy (deviation of predictions from ground truths) in model prediction. We account for such discrepancy in negative predictions using:
\begin{align}
    \label{eq:zeta}
    \zeta (y_i, z_i) = \quad
    \begin{cases}
        \hat{z}_i^n - \, \frac{1}{|\hat{z}_i^+|} \, \sum \limits_{p} \, \hat{z}_i^p, &\text{if } z_i \text{ is an optimal slot} \\
        0, &\text{otherwise}
    \end{cases}
\end{align}
where $\hat{z}_i^n \in \hat{z}_i^-$ and $\hat{z}_i^p \in \hat{z}_i^+$ inferred from ground-truths $y_i$. Accounting for the discrepancy in positive predictions is undesirable and we keep $\zeta (y_i, z_i) = 0$ for all $\hat{z}_i^p \in \hat{z}_i^+$. These three factors are incorporated in our proposed loss:
\begin{align}
    \label{eq:ordis}
    \mathcal{L}_{ordis} (y, z) = \Bigg[ \frac{1}{1 + exp \left( - \beta (z) \cdot \left( \frac{\zeta (y, z)}{2 + \delta(z) + \epsilon_2}  \right) 
 \right)} \Bigg] \cdot \mathcal{L}_f (y, z)
\end{align}
where $\epsilon_2$ is a constant and $\mathcal{L}_f (y, z)$ denotes the traditional focal loss~\cite{lin2017focal}. The sigmoidal factor we introduce in Eq.~\ref{eq:ordis} serves as a \textit{penalty actuator} (denoted by $\Omega$) and regulates the misclassification penalty of the existing focal loss for HOID.

\subsection{Training and inference}
\label{sec:train-test}
In the training phase, we follow the DETR-based ZS-HOID models~\cite{liao2022gen,ning2023hoiclip} and compute the Hungarian cost for matching the query slots to ground-truths. The loss to be minimized is then computed based on the matched pairs, which is composed of the human and object box regression loss ($\mathcal{L}_{b}$), generalized Intersection over Union (IoU) loss~\cite{rezatofighi2019generalized} ($\mathcal{L}_{u}$), object classification loss ($\mathcal{L}_{o}$), and the proposed ORDis loss for interaction classification:
\begin{align}
    \label{eq:total_loss}
   \mathcal{L} = \lambda_b \mathcal{L}_{b} + \lambda_u \mathcal{L}_{u} + \lambda_o \mathcal{L}_{o} + \mathcal{L}_{ordis}
\end{align}
where $\lambda_b, \lambda_u$, and $\lambda_o$ are weight coefficients for the corresponding losses. Given a test image during inference, the interaction embeddings produced by the decoder are post-processed, and interaction scores are obtained.

\section{Experiments}
\label{sec:expts}

\subsection{Datasets}
\label{sec:datasets}
We evaluate our model on two widely used benchmarks. The first is HICO-DET~\cite{chao2018learning}, which consists of 47,776 images in total. It contains 600 classes of HOI triplets, formed by combining $M = 117$ action classes and $N = 80$ object classes. Among these 600 interactions, 138 are considered Rare HOIs (each having less than 10 training instances), and the rest 462 HOI classes are considered Non-Rare. The second benchmark is V-COCO~\cite{gupta2015visual} containing 10,346 images annotated with 80 object classes and 29 action classes.

\begin{table}[t]
    \caption{Performance comparison on HICO-DET in the supervised setting (R = Rare, NR = Non-Rare).}
    \label{tab:supervised_hico}
    \centering
    \begin{tabular}{c|ccc|ccc}
    \toprule
    \multirow{2}{*}{\textbf{Method}}
    & \multicolumn{3}{c}{\textbf{Default}} 
    & \multicolumn{3}{c}{\textbf{Known Object}} \\
    & Full & R & NR & Full & R & NR \\
    \midrule
    \textbf{\textit{Two-stage}:} & & & & & & \\

    InteractNet~\cite{gkioxari2018detecting} & 9.94 & 7.16 & 10.77 &- &- &- \\

    iCAN~\cite{gao2018ican} & 14.84 & 10.45 & 16.15 & 16.26 & 11.33 & 17.73 \\

    iHOI~\cite{xu2019interact} & 13.39 & 9.51 & 14.55 &- &- &- \\

    DCA~\cite{wang2019deep} & 16.24 & 11.16 & 17.75 & 17.73 & 12.78 & 19.21 \\

    HRNet~\cite{gao2021hierarchical} & 18.10 &15.89 &18.76 &21.12 &18.20 &21.99 \\
    
    VCL~\cite{hou2020visual} &19.43 &16.55 &20.29 &22.00 &19.09 &22.87 \\

    RCD~\cite{9335510} &20.93 &18.95 &21.32 &23.02 &20.96 &23.42 \\

    IPGN~\cite{wang2021ipgn} &21.26 &18.47 &22.07 &- &- &- \\

    ACP++~\cite{kim2021acp++} &21.27 &15.41 &23.02 &25.61 &18.93 &27.6 \\

    ATL~\cite{hou2021affordance} &23.81 &17.43 &25.72 &27.38 &22.09 &28.96 \\ 

    DRG~\cite{gao2020drg} &24.53 &19.47 &26.04 &27.98 &23.11 &29.43 \\

    PPDM++~\cite{liao2024ppdm++} &31.20 &25.02 &33.05 &32.85 &26.18 &34.84 \\

    OSHOI~\cite{wu2024toward} &31.49 &28.23 &32.46 &- &- &- \\
    \midrule

    \textbf{\textit{One-stage}:} & & & & & & \\

    DSSF~\cite{gu2022dssf} &25.23 &18.72 &27.17 &28.53 &21.68 &30.57 \\
    
    QPIC~\cite{tamura2021qpic} &29.07 &21.85 &31.23 &31.68 &24.14 &33.93 \\

    AS-Net~\cite{chen2021reformulating} &28.87 &24.25 &30.25 &31.74 &27.07 &33.14 \\

    FGAHOI~\cite{ma2023fgahoi} &29.94 &22.24 &32.24 &32.48 &24.16 &34.97 \\ 

    CDT~\cite{zong2023zero} &30.48 &25.48 &32.37    &-  &- &- \\
    
    CDN~\cite{zhang2021mining} &31.78 &27.55 &33.05 &34.53 &29.73 &35.96 \\
    
    RMRQ~\cite{chan2024region} &31.11 &25.16 &32.88 &33.89 &27.78 &35.72 \\

    PDN~\cite{cheng2024parallel} & 31.84 &26.38 &33.47 &34.45 &29.34 &35.97 \\

    ERNet~\cite{lim2023ernet} &32.94 &27.86 &34.45 &- &- &- \\ 

    SG2HOI+~\cite{he2023toward} &32.62 &28.54 &35.17 &35.47 &31.36 &36.02 \\

   AFF~\cite{chan2024auxiliary} &33.08 &28.66 &34.41 &36.48 &32.79 &37.59 \\
   
    \midrule

    \textbf{Funnel-HOI} &\textbf{34.36} &\textbf{31.07} &\textbf{35.33} &\textbf{37.24} &\textbf{33.76} &\textbf{38.27} \\

    \bottomrule
    \end{tabular}

\end{table}

\subsection{Supervised and zero-shot splits}
\label{sec:zs_splits}

For V-COCO, there are 5,400 images in the trainval set and 4,946 in the test set which are used for presenting the model performance in a supervised setting. However, for this dataset, zero-shot splits are not defined in the current literature. 

The HICO-DET dataset is split into 38,118 images for training and 9,658 for testing, which is used for supervised model evaluation. In the zero-shot scenario, six different settings are defined in the existing literature. Specifically, under the Unseen Combination (UC) setting, examples from all the action and object categories are present, but a few combinations of them are unavailable during training. To this end,~\cite{bansal2020detecting} proposed 5 random splits of 480 seen and 120 unseen HOI classes. \cite{hou2020visual} proposed two special ways of assigning HOI classes as ``unseen'' -- by choosing either from the tail HOI classes (\textit{Rare First} or RF-UC) or head classes (Non-Rare First or NF-UC). The number of seen-unseen classes in both these cases is the same as in UC. \cite{bansal2020detecting} proposed Unseen Object (UO) as a more challenging setting where 12 objects are considered unseen during training, which results in 100 unseen and 500 seen HOIs. On a similar note,~\cite{liao2022gen} and~\cite{liu2020consnet} advocated the Unseen Verb (UV) and Unseen Action (UA) settings where 20 and 22 verbs, respectively, are unseen during training.

\subsection{Evaluation metrics}
\label{sec:metrics}
Following previous works, we choose the standard mean Average Precision (mAP) as our evaluation metric. 
For the V-COCO dataset, role AP is reported in two scenarios, where scenario 1 needs to predict the cases in which humans interact with no objects, while scenario 2 ignores these cases. On the other hand, we report performance in two settings for HICO-DET -- default and known object~\cite{chan2024auxiliary}. 
In the zero-shot scenario, mAP is reported for seen, unseen, and the full set.

\begin{figure*}[t]

  \centering
  \includegraphics[width=0.7\linewidth]{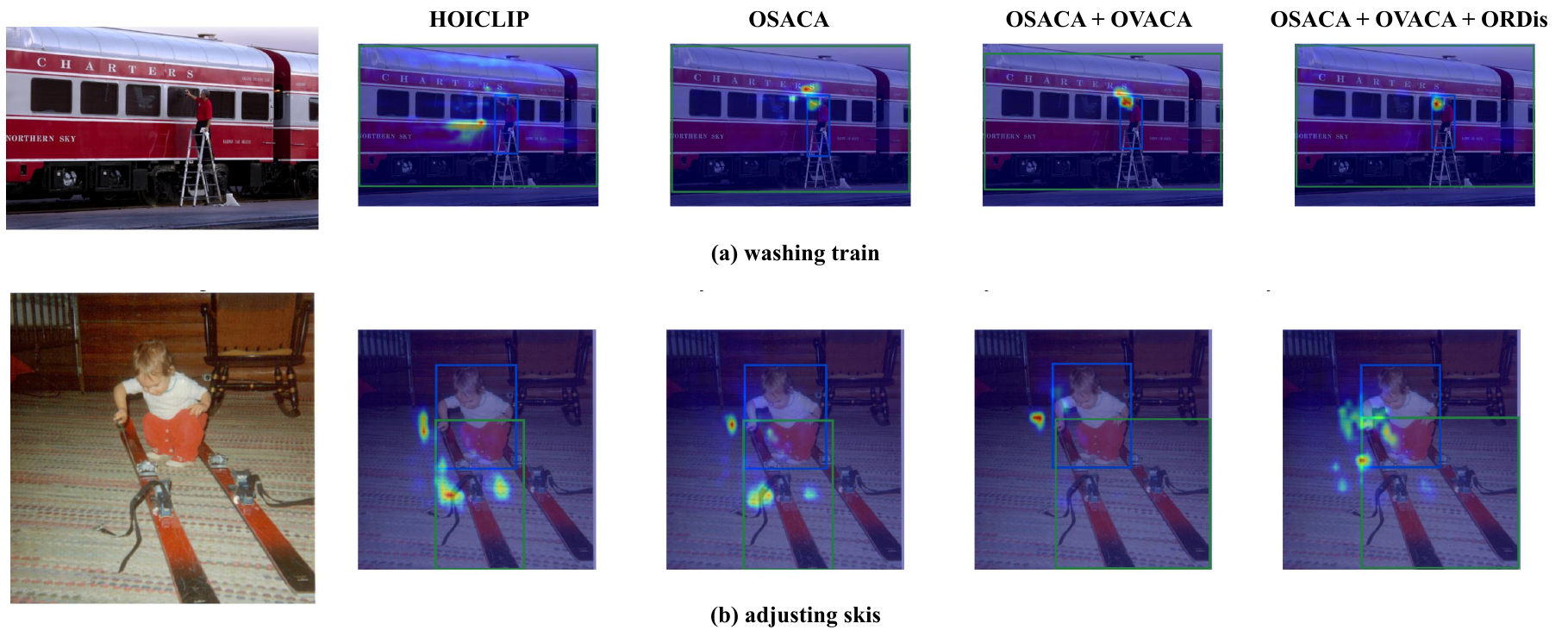}
  \caption{Attention visualization on the test images of HICO-DET for the current state-of-the-art method HOICLIP~\cite{ning2023hoiclip} as compared to Funnel-HOI (our method). Ground-truth interaction labels are shown below each case.}
  \label{fig:att-viz}
\end{figure*}

\subsection{Implementation details}
\label{sec:imple}
For a fair comparison, we use ResNet-50 as our CNN backbone, with $C_1 = 256$ channels for the visual feature map $\boldsymbol{\mathcal{V}_b}$. Meanwhile, we adopt the ViT-B/32 variant of CLIP that yields 512-dimensional embeddings ($d = 512$). The vanilla transformer encoder has 6 layers, while both the decoders have 3 layers each. The interaction decoder in our work follows~\cite{ning2023hoiclip}, along with the adoption of their verb adapter. The number of query slots $N_q$ is set to 64. For object/action nominations, $K_o$ and $K_a$ are both set to 5 and $C_2 = 64$ for both OSACA and OVACA. With regard to our loss function, we set $\kappa = 2, \epsilon_1 = 10^{-14}$, and $\epsilon_2 = 10^{-7}$. Following~\cite{liao2022gen}, the hyperparameters $\lambda_b, \lambda_u$, and $\lambda_o$ are set to 2.5, 1, and 1, respectively. During zero-shot training, the output dimension of the interaction classifier is set to the number of seen categories, while it is set to 600 during evaluation. Our model is trained for 90 epochs with a batch size of 8.
The learning rate is initially set to $10^{-4}$ and decays by a factor of 10 every 30 epochs. The overall framework is built on PyTorch using a single NVIDIA A100. 
Codes will be released on GitHub.

\begin{table}[t]
\caption{Performance comparison in V-COCO (fully-supervised setting). Average Precision is reported.}
\label{tab:v-coco}
\centering
\begin{tabular}{l|cc}
\hline
\textbf{Method} & $\boldsymbol{AP^{S1}_{role}}$ & $\boldsymbol{AP^{S2}_{role}}$ \\ \hline
iCAN~\cite{gao2018ican} & 45.3 & 52.4 \\
ConsNet~\cite{liu2020consnet} & 53.2   &  - \\
SCG~\cite{zhang2021spatially}          & 54.2   &   60.9 \\
QPIC~\cite{tamura2021qpic}          & 58.8    & 61.0   \\
CDN~\cite{zhang2021mining}  &  61.6  & 63.7   \\
GEN-VLKT~\cite{liao2022gen}         & \textbf{62.4}   &  64.4  \\

DSSF~\cite{gu2022dssf} & 52.7 &- \\

FGAHOI~\cite{ma2023fgahoi} &59.0	&59.3 \\

PPDM++~\cite{liao2024ppdm++} &54.3 &- \\

RMRQ~\cite{chan2024region} &61.05	&63.28 \\

Compo~\cite{zhuang2023compositional} & 57.2 & - \\

\midrule
\textbf{Funnel-HOI} & 61.6 & \textbf{64.6} \\
\bottomrule
\end{tabular}

\end{table}

\subsection{Zero-shot results}
\label{sec:results}
\subsubsection{Quantitative analysis}
\label{sec:quantitative}
Table~\ref{tab:results} shows that our method mostly achieves a higher mAP in all six settings. For UC, we report the mAP mean and standard deviation across five random splits, showing we achieve a relative mAP gain of 6.47\%, 4.93\%, and 5.18\% in unseen, seen, and full, respectively. For RF-UC, we obtain a relative mAP gain of 1.5\% on unseen and second-best results to HOICLIP~\cite{ning2023hoiclip} on seen and full. On the other hand, our seen and full results are best for NF-UC, but unseen is second-best to EoID. However, in the challenging UO, UA, and UV settings where objects and verbs can also be unseen, we achieve significant improvements over the state-of-the-art across unseen, seen, and full. Overall,  our method performs notably well on unseen classes (up to 12.4\% relative gain in UO) and also generalizes well in the \textit{full} scenario. These improvements can be attributed to the novel co-attention modules, which probe for multimodal object and action cues in a top-down manner, mimicking the human tendency to recognize unseen concepts while retaining generalizability.  

\begin{table}[t]
    \caption{Analysis of each component in our framework. \textit{OVACA} and \textit{OVACA + ORDis} are ineligible settings since OVACA always depends on the output of OSACA. }
    \label{tab:comp-ana}
    \centering
    \begin{tabular}{ccc|ccc}
    \toprule
    \textbf{OSACA} & \textbf{OVACA} & \textbf{ORDis} & \textbf{Unseen} & \textbf{Seen} & \textbf{Full}  \\
    \midrule 
    $\checkmark$ & & & \textbf{26.80} & 27.24 & 27.15 \\
    $\checkmark$ & $\checkmark$ & & 26.02 & 28.07 & 27.66 \\
    $\checkmark$ &  & $\checkmark$ & 26.14 & 28.13 & 27.73 \\
    &  & $\checkmark$ & 26.71 & 28.03 & 27.77 \\
    \midrule
    $\checkmark$ &  $\checkmark$ & $\checkmark$ & 26.63 & \textbf{28.32} & \textbf{28.00} \\
    \bottomrule
    \end{tabular}

\end{table}

\subsubsection{Qualitative analysis}
\label{sec:qualitative}
Figure~\ref{fig:att-viz} shows the attention visualization of HOICLIP predictions as compared to ours. We observe that the proposed components of our model progressively improve in discovering the region of interest for interaction cues (Fig.~\ref{fig:att-viz}a). Moreover, our detection is better than HOICLIP in some cases, such as in Fig.~\ref{fig:att-viz}b, where we detect both the \textit{skis}, unlike HOICLIP. Although OSACA could not detect both and emphasizes regions outside the detection box like HOICLIP, OVACA assists in detecting both of them. Along with our ORDis loss, our model finally shifts its focus to the interaction regions inside the detection boxes.

\begin{figure*}[t]

  \centering
  \includegraphics[width=\linewidth]{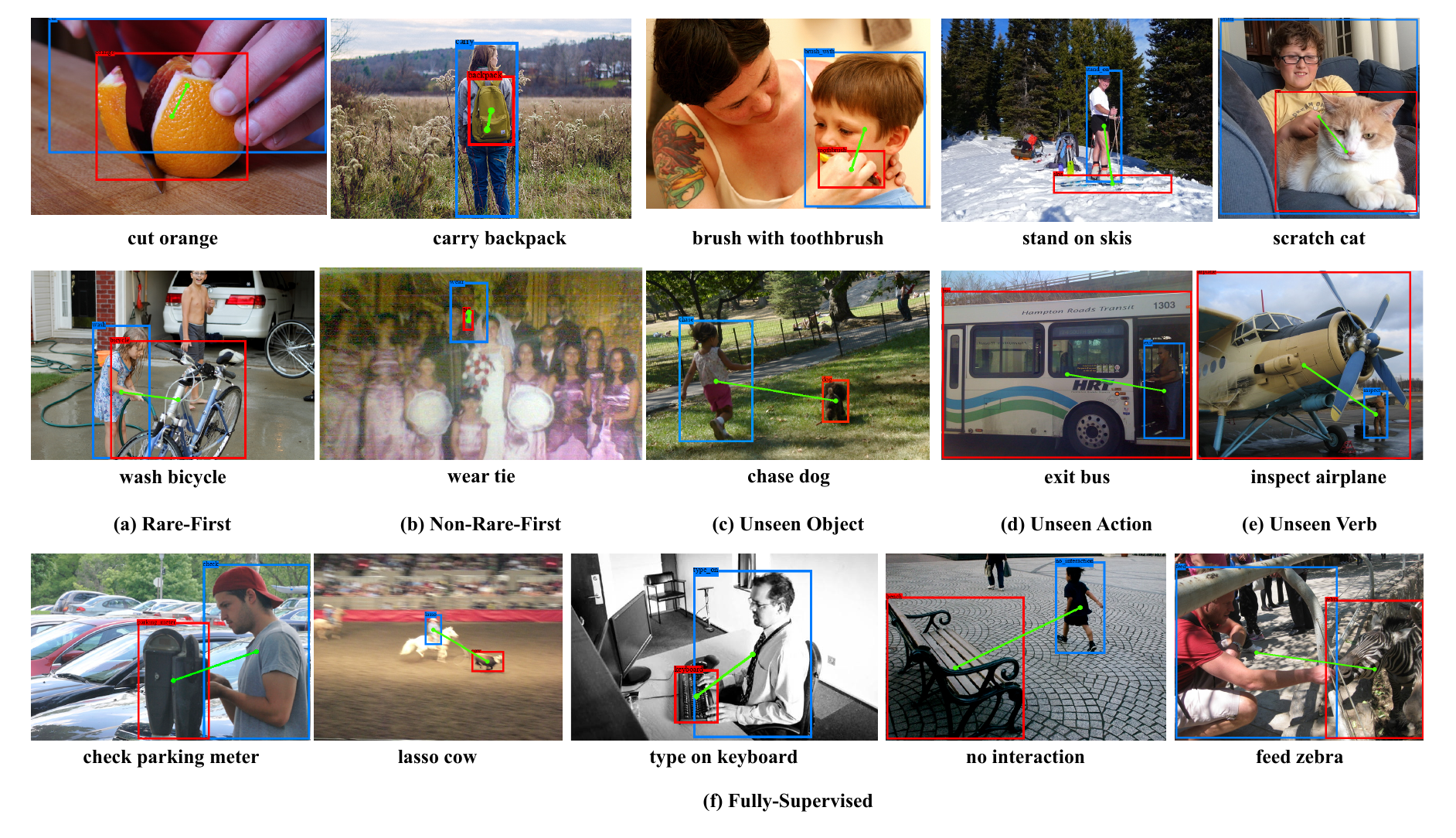}
  \caption{HOI detection results of our model on the HICO-DET test set in zero-shot (top two rows) and fully-supervised (bottom row) settings.}
  \label{fig:det-results}
\end{figure*}

Figures~\ref{fig:det-results}a to~\ref{fig:det-results}e demonstrate some insightful scenarios establishing the robustness of our model. For instance, \textit{cut orange} and \textit{wash bicycle} (Fig.~\ref{fig:det-results}a) are very rare interactions, with only 7 images each in HICO-DET. Small objects like \textit{toothbrush} and \textit{tie} are detected, and the interaction \textit{wear tie} (Fig.~\ref{fig:det-results}b) is captured even in a degraded image. Finally, interactions with partially occluded objects like \textit{skis} (Fig.~\ref{fig:det-results}d) are also correctly detected. 

\begin{table}[t]
    \caption{Analysis of each factor introduced in our ORDis loss.}
    \label{tab:ordis-ana}
    \centering
    \begin{tabular}{ccc|ccc}
    \toprule
    $\beta$ & $\delta$ & $\zeta$ & \textbf{Unseen} & \textbf{Seen} & \textbf{Full}  \\
    \midrule 
    $\checkmark$ & & & 26.12 & 27.30 & 27.07 \\
    & $\checkmark$ & & 26.37 & 27.58 & 27.40 \\
    & & $\checkmark$ & 26.21 & 27.74 & 27.40 \\
    $\checkmark$ & $\checkmark$ & & 26.50 & 27.96 & 27.67 \\
    $\checkmark$ & & $\checkmark$ & 26.60 & 27.82 & 27.57 \\
    & $\checkmark$ & $\checkmark$ & 26.56 & 28.06 & 27.76 \\
    \midrule
    $\checkmark$ &  $\checkmark$ & $\checkmark$ & \textbf{26.63} & \textbf{28.32} & \textbf{28.00} \\

    \bottomrule
    \end{tabular}

\end{table}

\begin{table}[t]
    \caption{Impact of $K_o$ on mAP by fixing $K_a = 5$.}
    \label{tab:ko}
    \centering
    \begin{tabular}{c|ccc}
    \toprule
    $\boldsymbol{K_o}$ &\textbf{Unseen} &\textbf{Seen} &\textbf{Full}  \\
    \midrule 
        1 &25.77 &\textbf{28.63} &28.06 \\
        3 &26.25 &28.40 &\textbf{29.97} \\
        5 &\textbf{26.63} &28.32 &28.00 \\
        7 &26.11 &28.05 &27.66 \\    
    \bottomrule
    \end{tabular}
\end{table}
\begin{table}[t]
    \caption{Impact of $K_a$ on mAP by fixing $K_o = 5$.}
    \label{tab:ka}
    \centering
    \begin{tabular}{c|ccc}
    \toprule
    $\boldsymbol{K_a}$ &\textbf{Unseen} &\textbf{Seen} &\textbf{Full}  \\
    \midrule 
           1 &25.93 &27.65 &27.31 \\
           3 &26.40 &27.65 &27.39 \\
           5 &\textbf{26.63} &\textbf{28.32} &\textbf{28.00} \\
           7 &26.34 &27.51 & 27.28 \\  

    \bottomrule
    \end{tabular}
\end{table}

\subsection{Results in the fully-supervised setting}
\label{sec:sup-res}
For the sake of completeness, we perform experiments in a fully-supervised setting and compare our results with existing two-stage and one-stage frameworks. In Tab.~\ref{tab:supervised_hico}, we compare with several methods for HICO-DET and obtain a higher performance in both default and known-object evaluations, with up to 8.4\% increase for rare HOIs. Almost all compared methods use a ResNet-50 backbone (except a few recent ones~\cite{lim2023ernet,ma2023fgahoi} using more powerful backbones), ensuring fair comparison. Qualitative results in Fig.~\ref{fig:det-results}f show certain interesting cases -- for instance, interaction \textit{lasso cow} is correctly detected in a blurry image, and a corner case like ``no interaction'' is also handled.

With V-COCO, Tab.~\ref{tab:v-coco} shows that we achieve comparable performance. As expected, we achieve a higher AP in Scenario 2 than in Scenario 1 since Scenario 1 contains HOIs \textit{independent of objects}, which is in contrast to our idea of associating verbs to the objects present in a scene.

\subsection{Ablation study}
\label{sec:ablations}
We choose the NF-UC setting for the ablation studies. It is a crucial setting as the training data consists of rare HOIs only, giving a ZS-HOID model the least number of samples to train on among all the zero-shot settings. Hence, learning robust zero-shot models in this setting becomes challenging. 
\subsubsection{Component analysis}
\label{sec:comp-ana}
Our first ablation aims to demonstrate the benefits of the three novel components of Funnel-HOI. Table~\ref{tab:comp-ana} shows that training our decoders with encoder outputs enhanced only by OSACA (i.e., if $\boldsymbol{\mathcal{F}_i} = \boldsymbol{\mathcal{F}_o} + \boldsymbol{\mathcal{F}_e}$), unseen mAP is significant, but model generalizability suffers (27.15 mAP in \textit{full}). One reason behind this could be that the object-probed feature map $\boldsymbol{\mathcal{F}_o}$ provides only a partial perspective towards anticipating interactions, guiding the interaction encodings $\boldsymbol{\mathcal{F}_i}$ weakly. The verb-probed feature map acquired according to the object presence (Eq.~\ref{eq:enc_output}) captures a more complete scene perception within $\boldsymbol{\mathcal{F}_i}$ due to integrated object and verb-based knowledge, providing a +0.51 mAP gain. The ORDis loss proves to be a crucial addition to our framework, achieving a high \textit{full} mAP by itself. However, it seems more compatible when used with OSACA, as seen mAP improves. Finally, the three components are complementary and seem to perform best when used together, maintaining a good balance between unseen and seen mAPs.

\subsubsection{Significance of different factors in ORDis loss}
\label{sec:ordis-factors}
While formulating our ORDis loss in Sec.~\ref{sec:ordis}, we introduced three factors to account for object perception ($\beta$), object-verb relatedness ($\delta$), and inter-class HOI prediction discrepancy ($\zeta$). Table~\ref{tab:ordis-ana} demonstrates the benefits of successively incorporating these factors. We observe a few interesting trends. First, $\delta$ usually benefits \textit{unseen} performance, most likely due to its induction of global information about object-verb associativity when unseen HOI triplets are encountered at test time. Second, $\zeta$ generally benefits overall (\textit{full}) performance, indicating that penalizing the discrepancy in negative predictions is a desirable property for ZS-HOID.
$\beta$ seems to act as a confidence booster. Again, the best performance is achieved when all three factors are used together.

\subsubsection{Impact of $K_o$ and $K_a$}
\label{sec:ko-kv}
We study our model performance by varying the number of nominated objects ($K_o$) and actions ($K_a$). From Tab.~\ref{tab:ka}, it is evident that nominating 5 action classes per image produces the best result. However, the most suitable $K_o$ is debatable, although performance clearly degrades beyond $K_o = 5$ (Tab.~\ref{tab:ko}).

\subsubsection{Model efficiency analysis}
\label{sec:efficiency}
Although our focus is more on improving mAP, we analyzed our model efficiency and found a slight increase in parameters for Funnel-HOI (66.8M as compared to 66.1M of HOICLIP~\cite{ning2023hoiclip}). On the HICO-DET dataset, our inference speed (56.59 ms/image) is comparable to that of HOICLIP (55.52 ms/image).  Furthermore, our GFLOPs (88.49) are marginally higher than HOICLIP (88.23) due to co-attention-based modules introduced in our method. However, the relative gain in performance achieved (up to $\sim 13\%$) justifies this trade-off. 

\begin{figure}[t]
  \centering
  \includegraphics[width=0.5\linewidth]{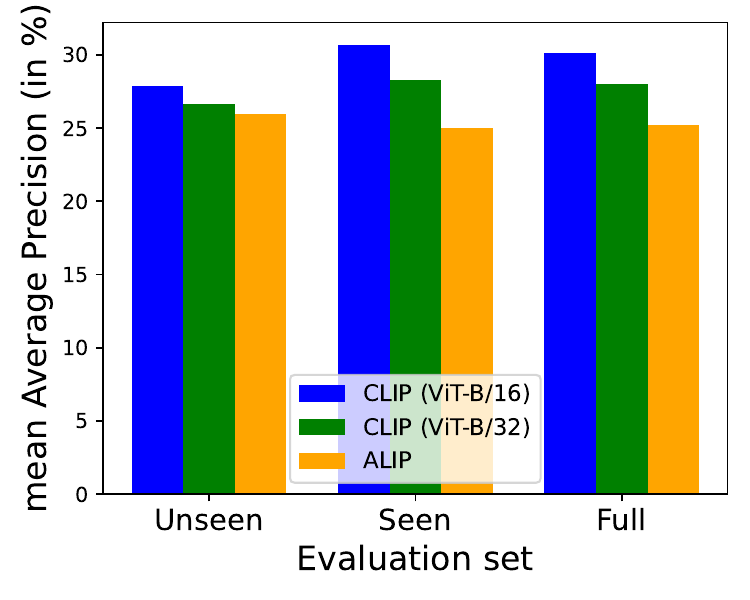}
  \caption{Performance of our model when coupled with different VLMs.}
  \label{fig:vlm}
\end{figure}

\subsubsection{Effect of pretrained vision-language models (VLMs)}
\label{sec:pretrained}
In our framework, we select CLIP (ViT-B/32) as a VLM for two reasons: (1) it is one of the very few publicly available VLMs, and (2) a fair performance comparison with other VLM-based ZS-HOID models is possible. To evaluate the impact of different VLM models used as feature extractors, we conduct two zero-shot experiments in the NF-UC setting with the HICO-DET dataset (summarized in Fig.~\ref{fig:vlm}). On employing a different CLIP variant (ViT-B/16), we witness a relative performance gain of 8\%. However, replacing CLIP with another recently-published VLM called ALIP~\cite{yang2023alip} results in a relative decline of 11\%. Therefore, multimodal frameworks for ZS-HOID are sensitive to representations from pretrained VLMs.

\section{Conclusion}
\label{sec:conclusion}
In this work, we propose Funnel-HOI -- a framework that focuses on improving ZS-HOID by replicating the human tendency to comprehend a scene in a top-down fashion. 
We observed that integrating multimodal object-action cues learned via a novel asymmetric co-attention mechanism fundamentally at the encoder level provides significant performance boosts. 
Additionally, our novel ORDis loss is well-aligned with the elements of HOI detection. We plan to extend our framework to handle HOIs in videos in the future.

\bibliographystyle{IEEEtran}
\bibliography{refs}

\begin{thebibliography}{10}
\providecommand{\url}[1]{#1}
\csname url@samestyle\endcsname
\providecommand{\newblock}{\relax}
\providecommand{\bibinfo}[2]{#2}
\providecommand{\BIBentrySTDinterwordspacing}{\spaceskip=0pt\relax}
\providecommand{\BIBentryALTinterwordstretchfactor}{4}
\providecommand{\BIBentryALTinterwordspacing}{\spaceskip=\fontdimen2\font plus
\BIBentryALTinterwordstretchfactor\fontdimen3\font minus
  \fontdimen4\font\relax}
\providecommand{\BIBforeignlanguage}[2]{{%
\expandafter\ifx\csname l@#1\endcsname\relax
\typeout{** WARNING: IEEEtran.bst: No hyphenation pattern has been}%
\typeout{** loaded for the language `#1'. Using the pattern for}%
\typeout{** the default language instead.}%
\else
\language=\csname l@#1\endcsname
\fi
#2}}
\providecommand{\BIBdecl}{\relax}
\BIBdecl

\bibitem{shen2018scaling}
L.~Shen, S.~Yeung, J.~Hoffman, G.~Mori, and L.~Fei-Fei, ``Scaling human-object
  interaction recognition through zero-shot learning,'' in \emph{2018 IEEE
  Winter Conference on Applications of Computer Vision (WACV)}.\hskip 1em plus
  0.5em minus 0.4em\relax IEEE, 2018, pp. 1568--1576.

\bibitem{carion2020end}
N.~Carion, F.~Massa, G.~Synnaeve, N.~Usunier, A.~Kirillov, and S.~Zagoruyko,
  ``End-to-end object detection with transformers,'' in \emph{European
  conference on computer vision}.\hskip 1em plus 0.5em minus 0.4em\relax
  Springer, 2020, pp. 213--229.

\bibitem{liao2022gen}
Y.~Liao, A.~Zhang, M.~Lu, Y.~Wang, X.~Li, and S.~Liu, ``Gen-vlkt: Simplify
  association and enhance interaction understanding for hoi detection,'' in
  \emph{Proceedings of the IEEE/CVF Conference on Computer Vision and Pattern
  Recognition}, 2022, pp. 20\,123--20\,132.

\bibitem{wu2023end}
M.~Wu, J.~Gu, Y.~Shen, M.~Lin, C.~Chen, and X.~Sun, ``End-to-end zero-shot hoi
  detection via vision and language knowledge distillation,'' in
  \emph{Proceedings of the AAAI Conference on Artificial Intelligence},
  vol.~37, no.~3, 2023, pp. 2839--2846.

\bibitem{ning2023hoiclip}
S.~Ning, L.~Qiu, Y.~Liu, and X.~He, ``Hoiclip: Efficient knowledge transfer for
  hoi detection with vision-language models,'' in \emph{Proceedings of the
  IEEE/CVF Conference on Computer Vision and Pattern Recognition}, 2023, pp.
  23\,507--23\,517.

\bibitem{radford2021learning}
A.~Radford, J.~W. Kim, C.~Hallacy, A.~Ramesh, G.~Goh, S.~Agarwal, G.~Sastry,
  A.~Askell, P.~Mishkin, J.~Clark \emph{et~al.}, ``Learning transferable visual
  models from natural language supervision,'' in \emph{International conference
  on machine learning}.\hskip 1em plus 0.5em minus 0.4em\relax PMLR, 2021, pp.
  8748--8763.

\bibitem{lin2017focal}
T.-Y. Lin, P.~Goyal, R.~Girshick, K.~He, and P.~Doll{\'a}r, ``Focal loss for
  dense object detection,'' in \emph{Proceedings of the IEEE international
  conference on computer vision}, 2017, pp. 2980--2988.

\bibitem{schubotz2014objects}
R.~I. Schubotz, M.~F. Wurm, M.~K. Wittmann, and D.~Y. von Cramon, ``Objects
  tell us what action we can expect: dissociating brain areas for retrieval and
  exploitation of action knowledge during action observation in fmri,''
  \emph{Frontiers in psychology}, vol.~5, p. 636, 2014.

\bibitem{gaspelin2018top}
N.~Gaspelin and S.~J. Luck, ``Top-down” does not mean “voluntary,''
  \emph{Journal of cognition}, vol.~1, no.~1, 2018.

\bibitem{chao2018learning}
Y.-W. Chao, Y.~Liu, X.~Liu, H.~Zeng, and J.~Deng, ``Learning to detect
  human-object interactions,'' in \emph{2018 ieee winter conference on
  applications of computer vision (wacv)}.\hskip 1em plus 0.5em minus
  0.4em\relax IEEE, 2018, pp. 381--389.

\bibitem{gupta2015visual}
S.~Gupta and J.~Malik, ``Visual semantic role labeling,'' \emph{arXiv preprint
  arXiv:1505.04474}, 2015.

\bibitem{gkioxari2018detecting}
G.~Gkioxari, R.~Girshick, P.~Doll{\'a}r, and K.~He, ``Detecting and recognizing
  human-object interactions,'' in \emph{Proceedings of the IEEE conference on
  computer vision and pattern recognition}, 2018, pp. 8359--8367.

\bibitem{gao2018ican}
C.~Gao, Y.~Zou, and J.-B. Huang, ``ican: Instance-centric attention network for
  human-object interaction detection,'' in \emph{British Machine Vision
  Conference}, 2018.

\bibitem{xu2019interact}
B.~Xu, J.~Li, Y.~Wong, Q.~Zhao, and M.~S. Kankanhalli, ``Interact as you
  intend: Intention-driven human-object interaction detection,'' \emph{IEEE
  Transactions on Multimedia}, vol.~22, no.~6, pp. 1423--1432, 2019.

\bibitem{hou2020visual}
Z.~Hou, X.~Peng, Y.~Qiao, and D.~Tao, ``Visual compositional learning for
  human-object interaction detection,'' in \emph{Computer Vision--ECCV 2020:
  16th European Conference, Glasgow, UK, August 23--28, 2020, Proceedings, Part
  XV 16}.\hskip 1em plus 0.5em minus 0.4em\relax Springer, 2020, pp. 584--600.

\bibitem{gao2020drg}
C.~Gao, J.~Xu, Y.~Zou, and J.-B. Huang, ``Drg: Dual relation graph for
  human-object interaction detection,'' in \emph{Computer Vision--ECCV 2020:
  16th European Conference, Glasgow, UK, August 23--28, 2020, Proceedings, Part
  XII 16}.\hskip 1em plus 0.5em minus 0.4em\relax Springer, 2020, pp. 696--712.

\bibitem{hou2021affordance}
Z.~Hou, B.~Yu, Y.~Qiao, X.~Peng, and D.~Tao, ``Affordance transfer learning for
  human-object interaction detection,'' in \emph{Proceedings of the IEEE/CVF
  Conference on Computer Vision and Pattern Recognition}, 2021, pp. 495--504.

\bibitem{gao2021hierarchical}
Y.~Gao, Z.~Kuang, G.~Li, W.~Zhang, and L.~Lin, ``Hierarchical reasoning network
  for human-object interaction detection,'' \emph{IEEE Transactions on Image
  Processing}, vol.~30, pp. 8306--8317, 2021.

\bibitem{wang2021ipgn}
H.~Wang, L.~Jiao, F.~Liu, L.~Li, X.~Liu, D.~Ji, and W.~Gan, ``Ipgn:
  Interactiveness proposal graph network for human-object interaction
  detection,'' \emph{IEEE Transactions on Image Processing}, vol.~30, pp.
  6583--6593, 2021.

\bibitem{liao2024ppdm++}
Y.~Liao, S.~Liu, Y.~Gao, A.~Zhang, Z.~Li, F.~Wang, and B.~Li, ``Ppdm++:
  Parallel point detection and matching for fast and accurate hoi detection,''
  \emph{IEEE Transactions on Pattern Analysis and Machine Intelligence}, 2024.

\bibitem{wu2024toward}
M.~Wu, Y.~Liu, J.~Ji, X.~Sun, and R.~Ji, ``Toward open-set human object
  interaction detection,'' in \emph{Proceedings of the AAAI Conference on
  Artificial Intelligence}, vol.~38, no.~6, 2024, pp. 6066--6073.

\bibitem{ren2015faster}
S.~Ren, K.~He, R.~Girshick, and J.~Sun, ``Faster r-cnn: Towards real-time
  object detection with region proposal networks,'' \emph{Advances in neural
  information processing systems}, vol.~28, 2015.

\bibitem{kim2021acp++}
D.-J. Kim, X.~Sun, J.~Choi, S.~Lin, and I.~S. Kweon, ``Acp++: Action
  co-occurrence priors for human-object interaction detection,'' \emph{IEEE
  Transactions on Image Processing}, vol.~30, pp. 9150--9163, 2021.

\bibitem{tamura2021qpic}
M.~Tamura, H.~Ohashi, and T.~Yoshinaga, ``Qpic: Query-based pairwise
  human-object interaction detection with image-wide contextual information,''
  in \emph{Proceedings of the IEEE/CVF Conference on Computer Vision and
  Pattern Recognition}, 2021, pp. 10\,410--10\,419.

\bibitem{zhang2021mining}
A.~Zhang, Y.~Liao, S.~Liu, M.~Lu, Y.~Wang, C.~Gao, and X.~Li, ``Mining the
  benefits of two-stage and one-stage hoi detection,'' \emph{Advances in Neural
  Information Processing Systems}, vol.~34, pp. 17\,209--17\,220, 2021.

\bibitem{zhang2022efficient}
F.~Z. Zhang, D.~Campbell, and S.~Gould, ``Efficient two-stage detection of
  human-object interactions with a novel unary-pairwise transformer,'' in
  \emph{Proceedings of the IEEE/CVF Conference on Computer Vision and Pattern
  Recognition}, 2022, pp. 20\,104--20\,112.

\bibitem{ma2023fgahoi}
S.~Ma, Y.~Wang, S.~Wang, and Y.~Wei, ``Fgahoi: Fine-grained anchors for
  human-object interaction detection,'' \emph{IEEE Transactions on Pattern
  Analysis and Machine Intelligence}, 2023.

\bibitem{zong2023zero}
D.~Zong and S.~Sun, ``Zero-shot human--object interaction detection via
  similarity propagation,'' \emph{IEEE Transactions on Neural Networks and
  Learning Systems}, 2023.

\bibitem{chan2024region}
S.~Chan, W.~Wang, Z.~Shao, Z.~Wang, and C.~Bai, ``Region mining and refined
  query improved hoi detection in transformer,'' \emph{IEEE Transactions on
  Emerging Topics in Computational Intelligence}, 2024.

\bibitem{cheng2024parallel}
Y.~Cheng, H.~Duan, C.~Wang, and Z.~Chen, ``Parallel disentangling network for
  human--object interaction detection,'' \emph{Pattern Recognition}, vol. 146,
  p. 110021, 2024.

\bibitem{bansal2020detecting}
A.~Bansal, S.~S. Rambhatla, A.~Shrivastava, and R.~Chellappa, ``Detecting
  human-object interactions via functional generalization,'' in
  \emph{Proceedings of the AAAI Conference on Artificial Intelligence},
  vol.~34, no.~07, 2020, pp. 10\,460--10\,469.

\bibitem{liu2020consnet}
Y.~Liu, J.~Yuan, and C.~W. Chen, ``Consnet: Learning consistency graph for
  zero-shot human-object interaction detection,'' in \emph{Proceedings of the
  28th ACM International Conference on Multimedia}, 2020, pp. 4235--4243.

\bibitem{hou2021detecting}
Z.~Hou, B.~Yu, Y.~Qiao, X.~Peng, and D.~Tao, ``Detecting human-object
  interaction via fabricated compositional learning,'' in \emph{Proceedings of
  the IEEE/CVF Conference on Computer Vision and Pattern Recognition}, 2021,
  pp. 14\,646--14\,655.

\bibitem{hou2022discovering}
Z.~Hou, B.~Yu, and D.~Tao, ``Discovering human-object interaction concepts via
  self-compositional learning,'' in \emph{European Conference on Computer
  Vision}.\hskip 1em plus 0.5em minus 0.4em\relax Springer, 2022, pp. 461--478.

\bibitem{sarullo2020zero}
A.~Sarullo and T.~Mu, ``Zero-shot human-object interaction recognition via
  affordance graphs,'' \emph{arXiv preprint arXiv:2009.01039}, 2020.

\bibitem{li-etal-2018-co}
\BIBentryALTinterwordspacing
X.~Li, K.~Song, S.~Feng, D.~Wang, and Y.~Zhang, ``A co-attention neural network
  model for emotion cause analysis with emotional context awareness,'' in
  \emph{Proceedings of the 2018 Conference on Empirical Methods in Natural
  Language Processing}, E.~Riloff, D.~Chiang, J.~Hockenmaier, and J.~Tsujii,
  Eds.\hskip 1em plus 0.5em minus 0.4em\relax Brussels, Belgium: Association
  for Computational Linguistics, Oct.-Nov. 2018, pp. 4752--4757. [Online].
  Available: \url{https://aclanthology.org/D18-1506}
\BIBentrySTDinterwordspacing

\bibitem{lu2019see}
X.~Lu, W.~Wang, C.~Ma, J.~Shen, L.~Shao, and F.~Porikli, ``See more, know more:
  Unsupervised video object segmentation with co-attention siamese networks,''
  in \emph{Proceedings of the IEEE/CVF conference on computer vision and
  pattern recognition}, 2019, pp. 3623--3632.

\bibitem{zhong2020polysemy}
X.~Zhong, C.~Ding, X.~Qu, and D.~Tao, ``Polysemy deciphering network for
  human-object interaction detection,'' in \emph{Computer Vision--ECCV 2020:
  16th European Conference, Glasgow, UK, August 23--28, 2020, Proceedings, Part
  XX 16}.\hskip 1em plus 0.5em minus 0.4em\relax Springer, 2020, pp. 69--85.

\bibitem{kuhn1955hungarian}
H.~W. Kuhn, ``The hungarian method for the assignment problem,'' \emph{Naval
  research logistics quarterly}, vol.~2, no. 1-2, pp. 83--97, 1955.

\bibitem{rezatofighi2019generalized}
H.~Rezatofighi, N.~Tsoi, J.~Gwak, A.~Sadeghian, I.~Reid, and S.~Savarese,
  ``Generalized intersection over union: A metric and a loss for bounding box
  regression,'' in \emph{Proceedings of the IEEE/CVF conference on computer
  vision and pattern recognition}, 2019, pp. 658--666.

\bibitem{wang2019deep}
T.~Wang, R.~M. Anwer, M.~H. Khan, F.~S. Khan, Y.~Pang, L.~Shao, and
  J.~Laaksonen, ``Deep contextual attention for human-object interaction
  detection,'' in \emph{Proceedings of the IEEE/CVF International Conference on
  Computer Vision}, 2019, pp. 5694--5702.

\bibitem{9335510}
Y.-L. Li, X.~Liu, X.~Wu, X.~Huang, L.~Xu, and C.~Lu, ``Transferable
  interactiveness knowledge for human-object interaction detection,''
  \emph{IEEE Transactions on Pattern Analysis and Machine Intelligence},
  vol.~44, no.~7, pp. 3870--3882, 2022.

\bibitem{gu2022dssf}
D.~Gu, S.~Ma, and S.~Cai, ``Dssf: Dynamic semantic sampling and fusion for
  one-stage human--object interaction detection,'' \emph{IEEE Transactions on
  Instrumentation and Measurement}, vol.~71, pp. 1--13, 2022.

\bibitem{chen2021reformulating}
M.~Chen, Y.~Liao, S.~Liu, Z.~Chen, F.~Wang, and C.~Qian, ``Reformulating hoi
  detection as adaptive set prediction,'' in \emph{Proceedings of the IEEE/CVF
  Conference on Computer Vision and Pattern Recognition}, 2021, pp. 9004--9013.

\bibitem{lim2023ernet}
J.~Lim, V.~M. Baskaran, J.~M.-Y. Lim, K.~Wong, J.~See, and M.~Tistarelli,
  ``Ernet: An efficient and reliable human-object interaction detection
  network,'' \emph{IEEE Transactions on Image Processing}, vol.~32, pp.
  964--979, 2023.

\bibitem{he2023toward}
T.~He, L.~Gao, J.~Song, and Y.-F. Li, ``Toward a unified transformer-based
  framework for scene graph generation and human-object interaction
  detection,'' \emph{IEEE Transactions on Image Processing}, vol.~32, pp.
  6274--6288, 2023.

\bibitem{chan2024auxiliary}
S.~Chan, X.~Zeng, X.~Wang, J.~Hu, and C.~Bai, ``Auxiliary feature fusion and
  noise suppression for hoi detection,'' \emph{ACM Transactions on Multimedia
  Computing, Communications and Applications}, vol.~20, no.~10, pp. 1--18,
  2024.

\bibitem{zhang2021spatially}
F.~Z. Zhang, D.~Campbell, and S.~Gould, ``Spatially conditioned graphs for
  detecting human-object interactions,'' in \emph{Proceedings of the IEEE/CVF
  International Conference on Computer Vision}, 2021, pp. 13\,319--13\,327.

\bibitem{zhuang2023compositional}
Z.~Zhuang, R.~Qian, C.~Xie, and S.~Liang, ``Compositional learning in
  transformer-based human-object interaction detection,'' in \emph{2023 IEEE
  International Conference on Multimedia and Expo (ICME)}.\hskip 1em plus 0.5em
  minus 0.4em\relax IEEE, 2023, pp. 1038--1043.

\bibitem{yang2023alip}
K.~Yang, J.~Deng, X.~An, J.~Li, Z.~Feng, J.~Guo, J.~Yang, and T.~Liu, ``Alip:
  Adaptive language-image pre-training with synthetic caption,'' in
  \emph{Proceedings of the IEEE/CVF International Conference on Computer
  Vision}, 2023, pp. 2922--2931.

\end{thebibliography}

\vfill

\end{document}